\documentclass[conference]{IEEEtran}
\IEEEoverridecommandlockouts

\usepackage{cite}
\usepackage{amsmath,amssymb,amsfonts}
\usepackage{graphicx}
\usepackage{textcomp}
\usepackage{xcolor}

\usepackage[T1]{fontenc}
\usepackage[utf8]{inputenc}
\usepackage{graphicx}
\usepackage{booktabs}
\usepackage{amsmath}
\usepackage{amssymb}
\usepackage{amsthm}
\usepackage{hyperref}
\usepackage{float}
\usepackage{algorithm}
\usepackage{algpseudocode}
\usepackage{array}
\usepackage{multirow}
\usepackage{tabularx}
\usepackage{siunitx}

\usepackage{orcidlink}

\newtheorem{definition}{Definition}

\def\BibTeX{{\rm B\kern-.05em{\sc i\kern-.025em b}\kern-.08em
    T\kern-.1667em\lower.7ex\hbox{E}\kern-.125emX}}
\begin{document}

\title{Sector-Mean: Deterministic Initialization of K-Means Centroids via Angular Sector Partitioning
}

\author{
\IEEEauthorblockN{
Abhiyan Dhakal\,\orcidlink{0009-0003-3143-416X}
}
\IEEEauthorblockA{
Kathmandu University\\
Dhulikhel, Nepal\\
itsabhiyandhakal@gmail.com
}
\and
\IEEEauthorblockN{
Pranish Kafle\,\orcidlink{0009-0005-7195-1561}
}
\IEEEauthorblockA{
Kathmandu University\\
Dhulikhel, Nepal\\
pranishkafle1@gmail.com
}
\and
\IEEEauthorblockN{
Rajani Chulyadyo\,\orcidlink{0000-0001-5432-0569}
}
\IEEEauthorblockA{
Kathmandu University\\
Dhulikhel, Nepal\\
rajani.chulyadyo@ku.edu.np
}
}
\maketitle

\begin{abstract}

K-Means is one of the most widely used clustering algorithms, but its susceptibility to initial centroid selection remains a primary bottleneck for its convergence speed and clustering accuracy. This paper proposes Sector-Mean Initialization, a deterministic initialization strategy with O(N) time complexity that partitions the two-dimensional data space into angular sectors around the global centroid and initializes centroids using sector-wise means. We evaluate the method on established two-dimensional benchmarks (SIPU, Birch) and multiple real-world datasets, comparing against random, K-Means++, and Max-Min initialization under identical Lloyd iterations. The statistical analysis of Friedman’s test (p<0.05) and Nemenyi post-hoc comparison indicates that, while delivering equivalent clustering quality as K-Means++ and Max-Min, Sector-Mean offers significant computational efficiency. Experimental results show that Sector-Mean reduces the initialization time by 74.9\% and 59.8\% in comparison to K-Means++ and max-min, respectively. And, it yields the lowest average number of iterations, achieving approximately 5\% fewer iterations than K-Means++ and 16\% fewer than max-min. These results highlight that Sector-Mean initialization offers a deterministic and computationally efficient initialization strategy while preserving cluster quality.
\end{abstract}

\begin{IEEEkeywords}
K-Means Clustering, Deterministic Initialization, Sector-Mean, Angular Partitioning, Centroid Initialization.
\end{IEEEkeywords}

\section{Introduction}

Clustering is a fundamental task in unsupervised learning that aims to partition a set of data objects into multiple groups or clusters such that objects within the same cluster exhibit high intra-cluster similarity and low inter-cluster similarity~\cite{han_jiawei_nodate}. Lloyd's algorithm~\cite{lloyd_least_1982}, commonly referred to as K-Means algorithm~\cite{macqueen_methods_1967}, is a foundational and widely used partition-based clustering algorithm due to its conceptual simplicity, scalability and ease of implementation.

K-Means works by iteratively assigning each data point to its nearest centroid and then updating each centroid as the mean of the points assigned to it, repeating this process until convergence. Despite its popularity, the performance of K-Means heavily depends on the choice of initial centroids~\cite{arthur_k-means_2007}. Poor initialization of the initial centroids can lead to slow convergence, unstable clustering outcomes, and suboptimal solutions caused by local minima. Thus, centroid initialization has become a critical factor influencing both the efficiency and effectiveness of K-Means clustering.

Since the clustering quality depends heavily on the selection of the initial centroid, numerous methods have been developed and widely adopted to mitigate the initialization sensitivity. For example, probabilistic methods such as K-Means++~\cite{arthur_k-means_2007} improve the clustering performance by selecting well-separated initial centroids, and hence offering theoretical guarantees on clustering quality and computational efficiency. However, such probabilistic approaches are inherently stochastic, which can lead to variability across runs, reduced reproducibility, and additional computational overhead due to repeated distance computations~\cite{bahmani_scalable_2012,bachem_approximate_2016}. On the other hand, deterministic strategies such as hierarchical partitioning, histogram/grid aggregation and geometry‑aware heuristics deliver repeatability and single‑pass behavior at the cost of potential sensitivity to data geometry or projection quality~\cite{su_search_2007,celebi_deterministic_2012,gingles_histogram-based_nodate,redmond_method_2007}.

Motivated by the limitations of the existing centroid initialization strategies, we propose a novel deterministic initialization method for K-Means, Sector-Mean. This method is specifically designed for two-dimensional data and performs a geometric partitioning of the data space based on angular coordinates relative to the global mean. The space is divided into angular sectors, and the mean of each sector is selected as an initial centroid. This approach ensures comprehensive spatial coverage, eliminates the randomness of the traditional initialization techniques, and enhances runtime efficiency with O(N) time complexity.

Also, while multiple deterministic initialization algorithms have been proposed these algorithms differ in assumptions, computational cost, and clustering behavior, systematic evaluations across consistent experimental settings are limited. To demonstrate the effectiveness of our algorithm, it was evaluated alongside multiple widely used baseline algorithms within a common unified experimental setup, enabling fair and transparent comparison in terms of both computational efficiency and clustering quality.

The paper makes three main contributions.
First, we propose ``Sector-Mean,'' a novel deterministic centroid initialization strategy with O(N) time complexity that minimizes the initialization latency while also preserving the clustering quality. Second, we conduct a comprehensive comparative analysis of the algorithm against multiple widely adopted clustering methods such as Random, K-Means++, and Max-Min~\cite{gonzalez_clustering_1985}, across datasets of varying size, structure, and complexity in a common experimental setup. Third, we provide strong empirical evidence, using the statistical validation (Friedman's test and Nemenyi post-hoc analysis), demonstrating that our approach significantly reduces initialization time and the number of iterations without compromising clustering accuracy.

\section{Related Works}

Centroid initialization has been long recognized as one of the most influential factors influencing both the convergence speed and the final quality of the K-means algorithm. Poor initialization of centroids often leads to undesirable outcomes, increased iteration, and also degrades the overall quality of the clusters. To reduce the initialization sensitivity, multiple research studies has been done on designing more reliable seeding strategies that improved the clustering performance while minimizing the computational overhead. Existing approaches can be categorized into probabilistic sampling methods, deterministic distance-based heuristics, divisive partitioning strategies, and density-driven techniques, each of them offering different trade-offs between robustness, reproducibility, and computational complexity.

The K-Means++ algorithm \cite{arthur_k-means_2007} is one of the most widely used probabilistic sampling methods. It selects initial centroids sequentially with probability proportional to the squared distance from the previously chosen seeds. Due to its strong empirical performance, K-Means++ is being used as a standard for initialization. Several variants of this algorithm have been developed to improve the efficiency of the algorithm, including parallel or multi-pass variants such as K-Means$\parallel$~\cite{bahmani_scalable_2012} and MCMC-based approximations~\cite{bachem_approximate_2016,bachem_distributed_2017}, as they reduce the number of full data passes while retaining similar theoretical guarantees. Despite these advantages, probabilistic methods remain non-deterministic, resulting in variability across runs and reduced reproducibility. 
Additionally, K-Means++ can introduce latency for large datasets or resource constrained settings due to its higher time complexity of $O(N.k)$. 

To eliminate randomness and ensure higher reproducibility, deterministic initialization approaches have been actively explored. Classic distance-based heuristics such as Max-Min~\cite{gonzalez_clustering_1985} select centroids that maximize pairwise separation, improving coverage of the data space. While they are effective, they rely on repeated distance evaluations and thus the computational cost remains similar to the probabilistic approaches. Divisive partitioning approaches, including Var-Part and PCA-Part~\cite{su_search_2007,celebi_deterministic_2012}, recursively split the data using variance or principal component projections to produce consistent and reproducible initial centroids. Although these methods exhibit deterministic behavior, they often involve additional projection, sorting, or recursive operations that increase computational overhead and may be sensitive to data orientation or projection quality.

Histogram and grid-based approaches divide the data space into small bins or cells in a single pass and place centroids at the representative points of these bins~\cite{gingles_histogram-based_nodate,redmond_method_2007}. These methods are simple, fast, and work well for large or streaming datasets. However, because the grid structure is fixed, they may not adapt well to the actual shape of the data, which can reduce accuracy when clusters are irregular. Density-based methods, such as Mean-Shift and Density Peaks~\cite{comaniciu_mean_2002,cheng_mean_1995}, find cluster centers by locating areas where the data is most concentrated. These methods can handle clusters of any shape and are generally more robust. However, they require extra computations, such as neighborhood or kernel calculations, which increase the runtime. As a result, they are often slower and less practical for quick K-Means initialization.

Most recently, methods with deterministic initialization and low computational cost have gained attention. Algorithms such as DISCERN~\cite{hassani_discern_2021}, DK-Means~\cite{jothi_dk-means_2019}, and cK-Means~\cite{layeb_ck-means_2023} provide efficient initial seeding, particularly in high-dimensional domains such as document or gene expression clustering. These approaches typically rely on additional preprocessing steps or assumptions about the data structure, so they mostly perform well in target settings but may not perform well in general settings. Closely related to our work are geometry-based strategies that use spatial ordering or angular information to guide centroid placement, including pointer-ring and radial selection techniques~\cite{niu_initializing_2006,noauthor_initial_nodate}. Although these approaches utilize geometric properties of the data, they require complex selection procedures and also involve sorting that limits strict linear-time performance or straightforward vectorization. These approaches are the most direct predecessors to our proposed method. However, existing angular heuristics often rely on complex sorting mechanisms or lack direct vectorization. Sector-Mean is positioned within this deterministic aggregation family but distinguishes itself by extending geometry-aware angular partitioning to a simplified, vectorized single-pass initializer. It is specifically optimized to address the need for efficient, general-purpose initialization strategies for two-dimensional spatial data, ensuring \(O(N)\) complexity and high reproducibility.

\section{Proposed Algorithm}

In this section, we present our Sector-Mean algorithm for initializing cluster centroids for Lloyd's algorithm. This method partitions the input space into $k$ angular sectors relative to the dataset's global mean to ensure a representative and deterministic initial distribution of centroids.

\subsection{Mathematical Framework}

Let $X = \{ \mathbf{x}_1, \mathbf{x}_2, \dots, \mathbf{x}_N \}$ be a dataset of $N$ points in $\mathbb{R}^2$, where each point $\mathbf{x}_i = (x_{i,1}, x_{i,2})$. Let $k$ be the desired number of clusters.

\begin{definition}[Global Center]
The global geometric center, $\mathbf{c}$, of the dataset is defined as the mean of all data points:
\begin{equation}
    \mathbf{c} = \frac{1}{N} \sum_{i=1}^{N} \mathbf{x}_i
\end{equation}
\end{definition}

\begin{definition}[Angular Mapping]
For each point $\mathbf{x}_i$, we define the angular displacement $\theta_i \in (-\pi, \pi]$ relative to the global center $\mathbf{c}$:
\begin{equation}
    \theta_i = \operatorname{arctan2}(x_{i,2} - c_2, x_{i,1} - c_1)
\end{equation}
\end{definition}

\begin{definition}[Sector Assignment]
The dataset is partitioned into $k$ disjoint sectors $S_0, S_1, \dots, S_{k-1}$. A point $\mathbf{x}_i$ is assigned to sector $S_j$ based on its angle $\theta_i$:
\begin{equation}
    j = \left\lfloor \frac{\theta_i + \pi}{2\pi} \cdot k \right\rfloor
\end{equation}
where $j \in \{0, 1, \dots, k-1\}$. To handle the edge case where $\theta_i = \pi$, the index is clamped to $k-1$.
\end{definition}

\subsection{Step-by-Step Algorithmic Flow}

Here is the pseudocode of the Sector-Mean algorithm.

\subsubsection{Step 1: Compute Global Mean}
The algorithm first calculates the geometric center $\mathbf{c}$ of the entire dataset. This point serves as the origin for the polar coordinate transformation.

\subsubsection{Step 2: Calculate Angular Positions}
For every data point $x_i \in X$, the angle $\theta_i$ relative to $\mathbf{c}$ is computed. This step is $O(N)$, as it requires exactly one pass over the data.

\subsubsection{Step 3: Define Angular Sectors}
The total angular range of $2\pi$ radians is segmented into $k$ equal parts. This creates a partitioning of the 2D plane originating from the global mean.

\subsubsection{Step 4: Assign Points to Sectors}
Each point $x_i$ is mapped to its corresponding sector index $j \in \{0, \dots, k-1\}$. Mathematically, this mapping is expressed as:
\begin{equation}
    j = \min\left( \left\lfloor \frac{\theta_i + \pi}{2\pi} \cdot k \right\rfloor, \ k-1 \right)
\end{equation}

\subsubsection{Step 5: Compute Sector Centroids}
The initial cluster centroids $\mu_j$ are defined as the mean of all observations within each sector $S_j$. To handle edge cases where a sector contains no data points, the centroid defaults to the global mean:
\begin{equation}
    \mu_j = 
    \begin{cases} 
    \frac{1}{|S_j|} \sum_{x \in S_j} x & \text{if } |S_j| > 0 \\
    \mathbf{c} & \text{if } |S_j| = 0
    \end{cases}
\end{equation}

\subsection{Pseudocode Implementation}

Here is the pseudocode of the Sector-Mean algorithm.

\begin{algorithm}
\caption{Sector-wise Mean Initialization (SMI)}
\label{alg:smi}
\begin{algorithmic}[1]
\Require $X \in \mathbb{R}^{N \times 2}$ (Dataset), $k \in \mathbb{Z}^+$ (Clusters)
\Ensure $M \in \mathbb{R}^{k \times 2}$ (Initial Centroids)

\State $\mathbf{c} \leftarrow \frac{1}{N} \sum_{i=1}^{N} \mathbf{x}_i$ \Comment{Compute global mean}
\State $Sums \leftarrow \text{zeros}(k, 2)$
\State $Counts \leftarrow \text{zeros}(k)$

\For{$i = 1$ \textbf{to} $N$}
    \State $\theta_i \leftarrow \operatorname{arctan2}(x_{i,2} - c_2, x_{i,1} - c_1)$
    \State $j \leftarrow \lfloor \frac{\theta_i + \pi}{2\pi} \cdot k \rfloor$
    \State $j \leftarrow \min(j, k-1)$ \Comment{Handle boundary conditions}
    
    \State $Sums_j \leftarrow Sums_j + x_i$
    \State $Counts_j \leftarrow Counts_j + 1$
\EndFor
\For{$j = 0$ \textbf{to} $k-1$}
    \If{$Counts_j > 0$}
        \State $M_j \leftarrow Sums_j / Counts_j$
    \Else
        \State $M_j \leftarrow \mathbf{c}$ \Comment{Fallback for empty sectors}
    \EndIf
\EndFor

\State \Return $M$
\end{algorithmic}
\end{algorithm}

\section{Experimental Setup}

This section presents the experimental setup used to evaluate the proposed Sector-Mean initialization method. We provide the description of the computational environment, datasets, baseline algorithms, parameter settings, evaluation metrics, and statistical analysis procedures used in the experiment.

All the experiments were conducted locally in a controlled environment for the precise measurement of the initialization time and total runtime. Baseline algorithms, like K-Means and K-Means++ and Max-Min were implemented directly in Python~\cite{van_rossum_python_2009} without relying on external libraries for the unbiased comparison. The experimental stack consisted of NumPy~\cite{harris_array_2020} for numerical computation, Pandas~\cite{team_pandas-devpandas_2026} for aggregation and analysis of results, and Matplotlib~\cite{hunter_matplotlib_2007} for visualization. Scikit-learn utilities~\cite{pedregosa_scikit-learn_2011} were used for dataset handling and metric computation. All timing measurements were obtained using Python's built-in time module. The same software configuration and execution environment were maintained across all experiments.

\subsection{Datasets}

To evaluate performance across multiple structural regimes, the evaluation spans synthetic clustering benchmarks, large-scale synthetic spatial data, and real-world datasets. Table~\ref{tab:datasets} summarizes the 15 two-dimensional datasets used, including sample cardinality and the number of clusters employed during evaluation.

\begin{table}[h]
\centering
\caption{Summary of dataset characteristics.}
\label{tab:datasets}
\footnotesize
\setlength{\tabcolsep}{3pt}
\begin{tabular}{@{}p{0.34\columnwidth} p{0.24\columnwidth} r r@{}}
\toprule
Dataset & Type & $N$ (samples) & Eval.\ $k$ \\
\midrule
\multicolumn{4}{@{}l}{\textit{Synthetic Datasets}} \\
A Series (A1--A3) & SIPU & 3{,}000--7{,}500 & 20, 35, 50 \\
S Series (S1--S4) & SIPU & 5{,}000 & 15 \\
Birch Set (Birch1--Birch3) & Birch & 100{,}000 & 100 \\
\midrule
\multicolumn{4}{@{}l}{\textit{Real-World Spatial Datasets}} \\
UJIIndoorLoc & 2D Subset & 19{,}937 & 4 \\
California Housing & Lat--Long Subset & 20{,}640 & 4 \\
\midrule
\multicolumn{4}{@{}l}{\textit{Real-World Datasets (PCA mapped to 2D)}} \\
Iris & PCA $\rightarrow$ 2D & 150 & 3 \\
Wine & PCA $\rightarrow$ 2D & 178 & 3 \\
MNIST & PCA $\rightarrow$ 2D & 70{,}000 & 10 \\
\bottomrule
\end{tabular}
\end{table}

\subsubsection{Synthetic Datasets}

Synthetic dataset were sourced primarily from the SIPU repository~\cite{franti_k-means_2018}. This repository provides highly controlled two-dimensional point sets having varying sample cardinality and cluster overlap. Datasets from the SIPU repository such as A-series and S-series were used, as they are widely adopted in clustering research due to their precisely defined geometric structures. A-series include three datasets (A1--A3) with increasing sample sizes of 3000, 5250 and 7500 and their corresponding cluster counts of 20, 35 and 50 cluster counts respectively. Similarly, S-series comprise four datasets (S1--S4) with 5000 fixed sample size and cluster count of 15 each, with varying cluster overlap characteristics.
Additionally, we use the Birch family of synthetic spatial datasets~\cite{zhang_birch_1996} (Birch1--Birch3), each consisting of 100{,}000 points distributed across 100 clusters.

\subsubsection{Real World Datasets}

To capture real-world variability, we included spatial coordinate readings from the UJIIndoorLoc dataset~\cite{torres-sospedra_ujiindoorloc_2014}, with 19,937 samples spread across 4 clusters, and latitude-longitude pairs from the California Housing dataset~\cite{kelley_pace_sparse_1997}, which comprises 20,640 data points with cluster count of 4. In addition, ommon tabular benchmarks, Iris~\cite{fisher_use_1936}, Wine~\cite{aeberhard_comparative_1994}, and MNIST~\cite{deng_mnist_2012}, were projected into two dimensions using Principal Component Analysis (PCA)~\cite{mackiewicz_principal_1993}. This dimensionality reduction ensures a consistent 2D geometric domain for evaluating the deterministic initialization logic of the Sector-Mean algorithm.

\subsection{Clustering Protocol}

All experiments use K-Means as the underlying clustering algorithm. The proposed method and baseline algorithm differ only in the procedure for selecting the initial centroid. This ensures that the performance differences arise solely from initialization. Random initialization, K-Means++ initialization, and Max-Min initialization are considered as baseline initialization methods in this experiment.

\subsubsection{Random Initialization}

Random initialization serves as a universal baseline, where k initial centroids are sampled uniformly from the datasets. While Random Initialization is computationally efficient and a standard algorithm in many implementations, this method is highly susceptible to stochastic variability. The randomness often results in poor initialization leading to slow convergence and a high probability of reaching suboptimal local minima, mostly in dataset with complex or overlapping clusters.

\subsubsection{K-Means++ Initialization}

K-Means++ improves random selection using a distance-aware probabilistic strategy. In this method, the first centroid is chosen at random, while each subsequent centroid is selected with a probability proportional to its squared distance from the nearest already-chosen centroid. This strategy generally lowers the expected sum-of-squares error and helps the algorithm converge faster. In our implementation, K-Means++ is used in its standard stochastic form, and we control the random seed to ensure reproducibility across repeated runs.

\subsubsection{Max-Min Initialization}

Max-Min initialization is a distance-based strategy that selects the first centroid randomly and each subsequent centroid as the point farthest from all previously chosen centroids. This ensures well-dispersed initial centers, which can improve convergence for widely separated clusters. However, its reliance on extreme distances makes it sensitive to outliers and clusters with uneven densities. Comparing our method against Max-Min highlights performance relative to a fully deterministic, dispersion-focused initializer.

\section{Evaluation}

This section describes the methodology used to assess the performance of the proposed clustering initialization strategies. Both the quality of the resulting clusters and the computational efficiency of the algorithms across multiple datasets. Moreover, statistical significance testing is conducted to determine whether observed performance differences are consistent.

\subsection{Evaluation Metrics}

Clustering performance was evaluated using both internal optimization measures and external clustering-quality metrics.

\subsubsection{Inertia (Sum of Squared Errors, SSE)}
Inertia measures compactness within cluster by computing the total sum of squared distances of data points to their respective centroids~\cite{macqueen_methods_1967}. Lower inertia values mean more cohesive clusters and show a better optimization of the clustering objective~\cite{ahmed_k-means_2020}.

\subsubsection{Computational Efficiency}
Efficiency was calculated in terms of the time required for centroid initialization, the total execution time of the clustering algorithm, and the number of iterations required for convergence. These indicators indicate the practical computational overhead associated with each initialization strategy~\cite{ikotun_k-means-based_2021}.

\subsubsection{Silhouette Coefficient}
The silhouette score evaluates cluster quality by simultaneously accounting for intra-cluster compactness and inter-cluster separation, with values ranging from $-1$ to $1$~\cite{rousseeuw_silhouettes_1987}. Higher scores indicate more distinct and well-formed clusters. For scalability, the coefficient was computed on a randomly drawn subsample of 10,000 observations, balancing computational feasibility with statistical reliability.

\subsubsection{External Validation (ARI and NMI)}
When ground-truth labels were available, external validation was conducted using the Adjusted Rand Index (ARI)~\cite{hubert_comparing_1985} and Normalized Mutual Information (NMI)~\cite{van_der_hoef_understanding_2019}. Both metrics assess the similarity between predicted clusters and true class distributions. Values close to 1 indicate strong alignment with the ground truth. The results are shown in the appendix section.

\subsection{Statistical Significance Analysis}

To assess the robustness and generalizability of observed performance differences across datasets, non-parametric statistical testing was employed~\cite{demsar_statistical_2006}. The Friedman test was first applied to compare the performance of the initialization strategies. If the null hypothesis was rejected at a significance level of $\alpha = 0.05$, Nemenyi post-hoc tests were conducted for pairwise comparisons. Results were visualized using Critical Difference (CD) diagrams, which summarize the relative performance of methods through average ranks. For the ranking procedure, metrics with lower-is-better semantics (inertia, number of iterations, and execution time) and higher-is-better semantics (Silhouette coefficient, ARI, and NMI) were treated accordingly to ensure consistent interpretability.

\section{Results and Discussion}

\begin{table}[h]
\centering
\caption{Mean Initialization Times (ms) (Per-Dataset).}
\label{tab:init_times}
\footnotesize
\begin{tabularx}{\columnwidth}{@{}>{\raggedright\arraybackslash}Xcccc@{}}
\toprule
\textbf{Dataset} & \textbf{K-Means++} & \textbf{Max-Min} & \textbf{Random} & \textbf{Sector-Mean} \\ 
\midrule
\multicolumn{5}{@{}l}{\textit{A Series}} \\
A1 set & 1.077 & 0.729 & 0.386 & 0.218 \\
A2 set & 1.344 & 0.696 & 0.276 & 0.197 \\
A3 set & 2.121 & 1.024 & 0.284 & 0.208 \\
\midrule
\multicolumn{5}{@{}l}{\textit{S Series}} \\
S1 set & 0.671 & 0.405 & 0.262 & 0.189 \\
S2 set & 0.657 & 0.414 & 0.281 & 0.210 \\
S3 set & 0.738 & 0.438 & 0.295 & 0.215 \\
S4 set & 0.691 & 0.417 & 0.268 & 0.200 \\
\midrule
\multicolumn{5}{@{}l}{\textit{Birch Sets}} \\
Birch1 set & 54.146 & 28.733 & 1.300 & 2.282 \\
Birch2 set & 54.332 & 28.030 & 1.329 & 1.981 \\
Birch3 set & 52.808 & 27.400 & 1.295 & 2.147 \\
\midrule
\multicolumn{5}{@{}l}{\textit{Spatial Datasets}} \\
California Housing (Lat--Long) & 0.731 & 0.488 & 0.443 & 0.604 \\
UJIIndoorLoc (2D subset) & 0.694 & 0.462 & 0.465 & 0.535 \\
\midrule
\multicolumn{5}{@{}l}{\textit{Datasets PCA mapped to 2D}} \\
Iris & 0.532 & 0.362 & 0.165 & 0.068 \\
MNIST & 12.489 & 6.437 & 0.926 & 1.907 \\
Wine & 0.529 & 0.362 & 0.165 & 0.069 \\
\bottomrule
\end{tabularx}
\end{table}

\begin{table}[h]
\centering
\caption{Per-Dataset Iteration Counts.}
\label{tab:iteration_counts}
\footnotesize
\begin{tabularx}{\columnwidth}{@{}Xcccc@{}}
\toprule
\textbf{Dataset} & \textbf{K-Means++} & \textbf{Max-Min} & \textbf{Random} & \textbf{Sector-Mean} \\
\midrule
\multicolumn{5}{@{}l}{\textit{A Series}} \\
A1 set & 19.450 & \textbf{17.450} & 21.350 & 34.000 \\
A2 set & 20.150 & 17.050 & 27.000 & \textbf{16.000} \\
A3 set & 19.100 & \textbf{23.950} & 31.900 & 26.000 \\
\midrule
\multicolumn{5}{@{}l}{\textit{S Series}} \\
S1 set & 17.600 & 9.000 & 18.000 & \textbf{7.500} \\
S2 set & 18.400 & 17.350 & 29.000 & \textbf{10.000} \\
S3 set & 22.300 & 24.550 & 23.100 & \textbf{15.000} \\
S4 set & 34.400 & 40.150 & 55.750 & \textbf{21.000} \\
\midrule
\multicolumn{5}{@{}l}{\textit{Birch Sets}} \\
Birch1 set & 87.500 & 103.700 & 107.100 & \textbf{58.000} \\
Birch2 set & \textbf{42.950} & 45.050 & 51.000 & 44.000 \\
Birch3 set & 85.300 & \textbf{76.050} & 131.750 & 132.000 \\
\midrule
\multicolumn{5}{@{}l}{\textit{Spatial Datasets}} \\
California Housing (Lat--Long) & 8.400 & 10.650 & 10.650 & \textbf{5.000} \\
UJIIndoorLoc (2D subset) & 10.050 & \textbf{5.000} & 5.800 & 6.000 \\
\midrule
\multicolumn{5}{@{}l}{\textit{PCA Mapped (2D)}} \\
Iris & 5.000 & \textbf{4.900} & 6.300 & 8.000 \\
MNIST & 174.450 & 246.300 & 187.500 & \textbf{155.000} \\
Wine & \textbf{5.000} & 5.250 & 6.900 & 9.000 \\
\bottomrule
\end{tabularx}
\end{table}

\begin{table}[h]
\centering
\caption{Statistical comparison of initialization strategies using the Friedman test followed by the Nemenyi post-hoc test across 15 datasets ($\alpha=0.05$).}
\label{tab:stat_comparison}
\footnotesize
\begin{tabularx}{\columnwidth}{@{}l l c c >{\raggedright\arraybackslash}X@{}}
\toprule
\textbf{Metric} & \textbf{Algorithm} & \textbf{Rank} & \textbf{p-value} & \textbf{Outcome} \\
\midrule
\multirow{4}{*}{\textbf{Inertia}}
& Sector-Mean & \textbf{2.133} & -- & \textit{(Control)} \\
& K-Means++   & 2.267 & 0.992 & Statistical Tie \\
& Max-Min     & 2.067 & 0.999 & Statistical Tie \\
& Random      & 3.533 & 0.016 & Sig. Different \\
\midrule
\multirow{4}{*}{\textbf{Init Time}}
& Sector-Mean & \textbf{1.533} & -- & \textit{(Control)} \\
& K-Means++   & 4.000 & $<0.001$ & Sig. Different \\
& Max-Min     & 2.800 & 0.036 & Sig. Different \\
& Random      & 1.667 & 0.992 & Statistical Tie \\
\midrule
\multirow{4}{*}{\textbf{Iterations}}
& Sector-Mean & \textbf{2.067} & -- & \textit{(Control)} \\
& K-Means++   & 2.133 & 0.999 & Statistical Tie \\
& Max-Min     & 2.233 & 0.985 & Statistical Tie \\
& Random      & 3.567 & 0.008 & Sig. Different \\
\midrule
\multirow{4}{*}{\shortstack{\textbf{Silhouette}\\\textbf{Coefficient}}}
& Sector-Mean & 2.333 & -- & \textit{(Control)} \\
& K-Means++   & \textbf{2.000} & 0.942 & Statistical Tie \\
& Max-Min     & 3.600 & 0.999 & Statistical Tie \\
& Random      & 2.067 & 0.006 & Sig. Different \\
\bottomrule
\end{tabularx}
\end{table}

This section presents a comparative evaluation of Sector-Mean algorithm against the baselines. The trade-offs between computational efficiency and clustering quality are analyzed, which are also supported by non-parametric statistical validation.

\subsection{Computational Efficiency}

The proposed Sector-Mean method demonstrates a clear advantage in computational scalability. As shown in Table~\ref{tab:init_times}, Sector-Mean initializes in approximately 2ms on large-scale synthetic datasets such as the Birch series ($N=100,000$). In contrast, K-Means++ requires over 54ms, and Max-Min requires roughly 28ms, showing Sector-Mean algorithm's dominance on 2D space, especially for large-scale datasets. Overall, Sector-Mean reduces initialization time by an average of 74.9\% compared to K-Means++, and 59.8\% compared to Max-Min. This validates that the single-pass $O(N)$ sectoring approach effectively eliminates the overhead associated with the iterative distance computations required by both K-Means++ and Max-Min.

Furthermore, efficient initialization translates to faster convergence of the Lloyd's algorithm. Table~\ref{tab:iteration_counts} details the number of iterations required to reach convergence. Sector-Mean achieves the lowest iteration count in the majority of test cases. For instance, on the complex Birch1 dataset, Sector-Mean reduced the iteration count to 58.0, compared to 87.5 for K-Means++, 103.7 for Max-Min and 107.1 for Random. While Sector-Mean shows higher iteration counts on Birch3 (132.0 vs 85.3 for K-Means++ and 76.050 for Max-Min), its superior aggregate rank (see Table \ref{tab:stat_comparison}) demonstrates that it generally provides a more effective initialization across diverse geometric distributions than the alternatives

\subsection{Clustering Quality}
Efficiency is often a trade-off for quality, but the results indicate that Sector-Mean maintains high clustering performance. While K-Means++ is explicitly designed to minimize variance, Sector-Mean achieves competitive inertia values across the majority of benchmarks, indicating that angular partitioning effectively approximates the optimal density centers. Furthermore, regarding structural validity, Sector-Mean achieves a mean silhouette score (0.502) virtually identical to that of K-Means++ (0.501), and even yields superior separation on datasets with distinct spatial structures such as the S-series and UJIndoorLoc. This suggests that the proposed deterministic approach captures the underlying data geometry as robustly as expensive probabilistic sampling.

\subsection{Statistical Significance Analysis}

Across the diverse set of 15 datasets, the Friedman test revealed statistically significant performance difference among the proposed and the baselines algorithms for all three evaluation metrics (Inertia: $\chi^2_F(3)=13.00$, $p=0.0046$; Init Time: $\chi^2_F(3)=35.72$, $p<0.0001$; Iterations: $\chi^2_F(3)=13.87$, $p=0.0031$) rejecting the null hypothesis that all the algorithms perform equivalently. Following the statistical comparison framework recommended by J. Dem\v{s}ar~\cite{demsar_statistical_2006}, we conducted Nemenyi post-hoc pairwise comparisons using the critical difference of CD=1.211 (see appendix).

As shown in Table~\ref{tab:stat_comparison}, Sector-Mean achieves clustering quality statistically comparable to K-Means++ and Max-Min while significantly outperforming random. For initialization time, it obtains the best rank and is significantly faster than both K-Means++ and Max-Min. For convergence, it requires fewer iterations than random with no significant difference from the optimized baselines. Overall, the proposed method consistently provides competitive effectiveness with statistically significant improvements in computational efficiency across datasets.

\section{Conclusion}

This paper proposed Sector-Mean, a deterministic initialization algorithm specifically designed to optimize K-Means clustering for two-dimensional data. By partitioning the data space into angular sectors, the method achieves O(N) time complexity and eliminates the variability of probabilistic approaches. Experiments show that Sector-Mean algorithm significantly reduces initialization time, while achieving the lowest convergence iterations. Moreover, statistical analysis confirms that this significant gain in computational efficiency does not compromise clustering quality. These findings establish Sector-Mean as a highly efficient and reproducible alternative for 2D spatial clustering tasks.

\bibliographystyle{IEEEtran}
\bibliography{main}

\end{document}